\documentclass[sigconf,nonacm]{acmart}

\usepackage{algorithm}
\usepackage{algpseudocode}

\begin{document}

\raggedbottom

\title{Exploring the Privacy--Size Tradeoff for Differentially Private Self-Attentive Variational Autoencoders}

\author{Rohit Sadavarte}
\authornote{Corresponding author.}
\affiliation{%
  \institution{Foresight, Intuit}
  \city{Mountain View}
  \state{CA}
  \country{USA}}
\email{rohit\_sadavarte@intuit.com}

\author{Dhruvi Kothari}
\affiliation{%
  \institution{Foresight, Intuit}
  \city{Mountain View}
  \state{CA}
  \country{USA}}

\author{Sparsh Gupta}
\affiliation{%
  \institution{Foresight, Intuit}
  \city{Mountain View}
  \state{CA}
  \country{USA}}

\author{Eric Van Oeveren}
\affiliation{%
  \institution{Foresight, Intuit}
  \city{Mountain View}
  \state{CA}
  \country{USA}}

\begin{abstract}
Data tends to be most useful when it is shared, and the data most worth
sharing is often the data that privacy obligations restrict. Synthetic data
from a generator trained under differential privacy (DP) is one way to
reconcile the two, but its utility depends on the privacy budget
$\varepsilon$ and on how much data the generator sees - a parameter that most data sharing applications of synthetic data control.
We present SAVANT, a self-attentive variational autoencoder for tabular
data trained with DP-SGD that, to our knowledge, is the first to combine
attention, a VAE, and DP in one tabular model. We evaluate it against two
marginal mechanisms, PrivBayes and MST, jointly over the privacy budget and
the dataset size on two public datasets, measuring the utility a downstream
classifier loses by training on synthetic rather than real data. We find that, under the same DP guarantee, more data
generally means more utility. For the marginal mechanisms it absorbs most of
what a strict budget costs, and a noise-scaling argument suggests DP-SGD
exchanges data for budget at a worse rate. Two independent
membership-inference attacks, judged against a measured null, found no
aggregate membership signal.
\end{abstract}

\maketitle

\section{Introduction}

Differentially private (DP) synthetic data generation fits a generative model
to a sensitive table and releases samples from the model. Sampling from a fitted
model is not private by itself, since the model can memorize training rows, so
the fit is made under a formal guarantee~\cite{dwork2006calibrating}. The
guarantee is set by a privacy budget $\varepsilon$: noise injected during
fitting bounds how much the output can reveal about any one record, and a
smaller $\varepsilon$ means more noise and less utility. Mechanisms differ in
where that noise enters. Marginal and graphical mechanisms such as
PrivBayes~\cite{zhang2017privbayes} and MST~\cite{mckenna2021winning} perturb a
fixed set of low-order statistics once, while DP-SGD~\cite{abadi2016deep}
perturbs every gradient step of a deep generative model.

Utility depends on $\varepsilon$ and on the number of training rows
$n$, and existing evidence mostly addresses the privacy tradeoff. Evaluations of DP
synthetic tabular data usually sweep $\varepsilon$ at one fixed $n$, so they
cannot say whether the two trade against each other. The question matters
because $n$ is often a choice: retention windows, how many sources to pool, and
how long to collect before fitting are all settings someone sets. If the
utility lost to a strict budget shrinks as $n$ grows, a team held to a small
$\varepsilon$ has a second lever that raises no further privacy question. That
is the problem we faced with tabular synthetic data generation, and the one
we study here.

We measure the two axes jointly, for three mechanisms from two families: the
marginal mechanisms PrivBayes and MST, and SAVANT, a DP-SGD-trained generative
model. Our DP-SGD trained model, SAVANT, replaces the fully connected stacks of a tabular VAE with
self-attention over learned per-column embeddings, and computes its loss per
sample for the DP-SGD specific implementation. (\S\ref{sec:savant}). To our
knowledge no earlier tabular model combines attention, a VAE, and is trained for DP.
(\S\ref{sec:related}).

Since the setting is security-sensitive, we also audit privacy empirically
on the same grid rather than rely on the formal guarantee alone. A
distance-based attack and a density-ratio attack (DOMIAS) run independently on
every cell and are judged against a measured null, not against zero
(\S\ref{sec:audit}). In short, we show that data volume and budget are partial
substitutes for marginal mechanisms and that the exchange rate differs for
DP-SGD, we present a self-attentive DP VAE and what makes it trainable, and we
audit all three mechanisms against two independent attacks. The evidence comes
from one implementation per mechanism and two public datasets whose
trends matched our internal needs most closely; we present it as
empirical evidence, grounded in theory, for practical
data-volume-versus-budget decisions.

\section{Related Work}
\label{sec:related}

Benchmarks converge on marginal mechanisms outperforming deep generative ones
on downstream utility~\cite{chen2025benchmarking}, but sweep $\varepsilon$ at
one dataset size. Systematic benchmarks identify the best-performing
algorithms~\cite{tao2021benchmarking} and report a utility-efficiency
trade-off~\cite{chen2025benchmarking}, while MargNet~\cite{chen2025margnet} among other works
argues the consensus favoring statistical models is incomplete for densely
correlated data, where neural networks are better suited. These establish
which mechanism wins in a given setting, not how that ranking moves with
dataset size. The closest prior art is Ganev et
al.~\cite{ganev2023graphical}, who observe that graphical models improve
monotonically with data while deep generative models spend budget per training
iteration and behave less predictably. We agree with and quantify that
observation, measuring the interaction with $\varepsilon$ rather than the main
effect of data, on one grid.

Attention has been applied to tabular VAEs without
privacy~\cite{wang2025ttvae}. DP has been applied to attention models that are
not VAEs: self-attention GANs~\cite{li2023dpsactgan}, transformer GANs under
PATE~\cite{youssef2026patetabtransgan}, autoregressive
transformers~\cite{castellon2023dptbart}, and a latent diffusion model over a
DP-SGD autoencoder~\cite{zhu2024dptldm}. Vanilla DP VAEs have been implemented earlier. SAVANT occupies that gap. On privacy auditing,
we follow the view that synthetic data should be held to measured attack
results~\cite{stadler2022groundhog,houssiau2022tapas}, starting from membership
inference~\cite{shokri2017membership}, and we do not rely on distance-based
proxies alone~\cite{yao2025dcr,ganev2025inadequacy}, which is why we implement
two attacks with different assumptions (\S\ref{sec:audit}) as another emperical check against these attacks.

\section{SAVANT and Experimental Setup}
\label{sec:savant}

\subsection{SAVANT}

\paragraph{Architecture.}
SAVANT replaces the fully connected encoder and decoder of a tabular VAE with
multi-head self-attention blocks without positional encodings, since columns
carry no order. Categorical columns use learned embeddings of size
$\lceil\sqrt{\text{cardinality}}\rceil$ instead of one-hot vectors, which keeps
the attention parameter count independent of total vocabulary; this matters on
high-cardinality columns such as ACS-PUMS's \texttt{OCCP} ($\sim$500
categories), which resemble enterprise data. Continuous and discrete
features pass through separate projections, and the decoder mirrors the
encoder with a learned per-dimension $\sigma$. We also experimented with other
prior distributions in place of the Gaussian. The tunable hyperparameters are
the embedding and latent dimension, the encoder and decoder stack widths, the
number of attention heads, and dropout, together with the DP-SGD training
settings: epochs, batch size, learning rate, gradient clipping norm, and the
reconstruction weight.

\paragraph{Shaped for DP-SGD.}
The objective is a per-sample ELBO: a Gaussian likelihood with learned $\sigma$
for continuous columns, per-column cross-entropy for categorical columns, and
the closed-form KL term. The loss is returned per sample rather than reduced,
which DP-SGD requires to clip per-example gradients; a loss written only in
reduced form must be re-derived before it can be trained privately at all.

\subsection{Mechanism configurations}

\emph{PrivBayes} (DataSynthesizer): Bayesian network degree $k=1$, so each
column is conditioned on one parent; pure $\varepsilon$-DP via the Laplace
mechanism; category domains and continuous bounds supplied explicitly.
\emph{MST} (smartnoise-synth): default pairwise-marginal selection with
Private-PGM estimation; continuous columns discretized to 20 equal-width
bins over the supplied public bounds; $(\varepsilon,\delta)$-DP with
$\delta = 10^{-5}$; preprocessor budget set to zero so the entire
$\varepsilon$ reaches the mechanism. \emph{SAVANT}: with the
hyperparameters described in \S\ref{sec:savant}, trained with Opacus
($\geq$1.4, ghost clipping) at $\delta = 10^{-5}$.

\paragraph{Domain as public knowledge.}
Every mechanism receives column domains and continuous bounds as public
inputs at zero privacy cost. This equalizes the mechanisms, and is also a
practical necessity: smartnoise's private bounds inference fails at
$\varepsilon=0.1$, the corner of most interest here.

\subsection{Data and grid}
We use UCI Adult~\cite{kohavi1996scaling} (48{,}842 rows) and ACS-PUMS 2018
ACS-Income~\cite{ding2021retiring} pooled over 20 states, subsampled to
nested sizes of 10K, 25K, 50K, 250K, and 500K rows, so a size effect is never
confounded with sampling different rows. Adult has three sizes, up to
48.8K. We chose these two for this study because, they have a mix of numeric and high cardinality features, a trend common for some enterprise datasets. 
The grid is $\varepsilon \in
\{0.1,1,5,10,20\}$ with $\delta=10^{-5}$ where applicable. Column domains are
supplied as public knowledge so every mechanism spends its whole budget on
fitting. PrivBayes~\cite{zhang2017privbayes} (DataSynthesizer) and
MST~\cite{mckenna2021winning} (smartnoise-synth) are the baselines.

\paragraph{Splits and seeds.}

Size tiers are nested prefixes of a single seeded shuffle, so a smaller tier
is a subset of every larger one and a size effect cannot be confounded with
having drawn different rows. For each cell, a stratified 20\% test split is
reserved \emph{before} the synthesizer sees any data; the synthesizer is fit
on the remaining 80\%, and both the real baseline and the synthetic-trained
model are evaluated on that same held-out split.

\subsection{Metrics}
Our primary metric is \emph{utility drop}: the F1 a downstream classifier loses
by training on synthetic rather than real data, both evaluated on the same
held-out real split (logistic regression, binary income prediction; real-data
F1 is 0.656 on Adult and 0.744 on ACS). Subtracting the matched baseline makes
cells comparable across seeds and datasets. A task-independent check, the mean
correlation difference over all column pairs, is defined below.

Let $R_{\text{tr}}, R_{\text{te}}$ be the real train and test partitions and
$S$ the synthetic table, $|S| = |R_{\text{tr}}|$.

\paragraph{Utility drop.}
Fit logistic regression (scikit-learn defaults, one-hot categoricals,
standardized continuous features) on $R_{\text{tr}}$ and on $S$; score both
on $R_{\text{te}}$; report
$\mathrm{F1}(R_{\text{tr}}) - \mathrm{F1}(S)$. This tests the usability of the data, and is a metric used consistently across literature.

\paragraph{Marginal fidelity.}
Per continuous column, the two-sample Kolmogorov--Smirnov statistic between
$S$ and $R_{\text{tr}}$, averaged over columns. Per categorical column, the
total variation distance
$\tfrac{1}{2}\sum_{c} \lvert p_S(c) - p_R(c) \rvert$, averaged over columns.
Both sides are cast to a common string representation first; without this,
mechanisms that emit categories as strings and mechanisms that emit them as
floats are scored against disjoint category sets and the distance is
silently maximal.

\paragraph{Mean correlation difference.}
Expand every categorical column into indicator columns, keeping only the
categories that cover 95\% of real mass so that rare categories do not
dominate the average with sampling noise. Compute the Pearson correlation
matrix on the expanded frame for $R_{\text{tr}}$ and for $S$; the
coefficient is then the $\phi$ coefficient for two indicators and the
point-biserial correlation for an indicator against a continuous column.
Report the mean of $\lvert \rho_R(i,j) - \rho_S(i,j) \rvert$ over all pairs
$(i,j)$ that originate in \emph{different} source columns; indicator pairs
from the same source column are excluded, since they are related by
construction. Constant columns, whose correlation is undefined, contribute 0.

\paragraph{Quality.}
\emph{Category coverage}: the share of real categories that appear at least
once in $S$, averaged over categorical columns. \emph{Boundary adherence}: the share
of synthetic values inside the real bounds, which is 1.000 throughout because
bounds are supplied to every mechanism. \emph{Exact-match rate}: the share of
synthetic rows whose full stringified tuple equals some row of
$R_{\text{tr}}$.

\paragraph{Coincidence rate.}
The reference for the exact-match rate. Rather than compare against zero, we
measure how often a row of the held-out split $R_{\text{te}}$, consisting of
real people the synthesizer never saw, exactly matches a row of
$R_{\text{tr}}$ on the same schema. Any match rate below this floor is
attributable to population density on a coarse schema rather than to
memorization.

\begin{table*}[t]
  \caption{Utility drop (F1 lost by training on synthetic rather than real data; lower is better) on ACS-PUMS at the two ends of the budget range, and the penalty $\Delta$ for tightening from $\varepsilon=20$ to $0.1$. Seed means; 500K cells use 3 seeds, others 1.}
  \label{tab:headline}
  \small
  \begin{tabular}{r ccc ccc ccc}
    \toprule
    & \multicolumn{3}{c}{PrivBayes} & \multicolumn{3}{c}{MST} & \multicolumn{3}{c}{SAVANT} \\
    \cmidrule(lr){2-4}\cmidrule(lr){5-7}\cmidrule(lr){8-10}
    Rows & $\varepsilon{=}20$ & $0.1$ & $\Delta$ & $\varepsilon{=}20$ & $0.1$ & $\Delta$ & $\varepsilon{=}20$ & $0.1$ & $\Delta$ \\
    \midrule
    10K & 0.070 & 0.310 & +0.240 & 0.080 & 0.536 & +0.456 & 0.150 & 0.623 & +0.473 \\
    25K & 0.073 & 0.241 & +0.168 & 0.071 & 0.524 & +0.453 & 0.167 & 0.528 & +0.361 \\
    50K & 0.078 & 0.282 & +0.204 & 0.061 & 0.382 & +0.321 & 0.047 & 0.545 & +0.499 \\
    250K & 0.082 & 0.103 & +0.021 & 0.067 & 0.169 & +0.103 & 0.056 & 0.690 & +0.634 \\
    500K & 0.098 & 0.087 & -0.012 & 0.070 & 0.120 & +0.050 & 0.039 & 0.697 & +0.657 \\
    \bottomrule
  \end{tabular}
\end{table*}

\section{Results}

\paragraph{Budget cost is concentrated below $\varepsilon \approx 5$.}
Tightening from $\varepsilon=20$ to $5$ costs at most 0.07 F1 anywhere in the
study, while going to $\varepsilon=0.1$ costs up to 0.53. Budget above
$\varepsilon\approx 5$ buys almost no utility, so the decisions that matter are
at the strict end, and that is where data volume does its work.
Tables~\ref{tab:app-adult} and~\ref{tab:app-acs} give the utility drop for every
mechanism, budget, and size.

\begin{table}[h]
  \caption{Utility drop on Adult, seed means (3 seeds per cell).}
  \label{tab:app-adult}
  \small
  \begin{tabular}{llrrr}
    \toprule
    Mechanism & $\varepsilon$ & 10K & 25K & 48.8K \\
    \midrule
    PrivBayes & 0.1 & 0.615 & 0.571 & 0.495 \\
     & 1 & 0.553 & 0.495 & 0.364 \\
     & 5 & 0.411 & 0.397 & 0.288 \\
     & 10 & 0.390 & 0.376 & 0.285 \\
     & 20 & 0.313 & 0.305 & 0.279 \\
    \midrule
    MST & 0.1 & 0.440 & 0.164 & 0.190 \\
     & 1 & 0.212 & 0.229 & 0.197 \\
     & 5 & 0.211 & 0.207 & 0.174 \\
     & 10 & 0.218 & 0.209 & 0.176 \\
     & 20 & 0.219 & 0.207 & 0.168 \\
    \midrule
    SAVANT & 0.1 & 0.560 & 0.541 & 0.449 \\
     & 1 & 0.389 & 0.540 & 0.399 \\
     & 5 & 0.295 & 0.175 & 0.207 \\
     & 10 & 0.231 & 0.239 & 0.292 \\
     & 20 & 0.239 & 0.169 & 0.239 \\
    \bottomrule
  \end{tabular}
\end{table}

\begin{table}[h]
  \caption{Utility drop on ACS-PUMS, seed means (1 seed per cell, 3 at 500K).}
  \label{tab:app-acs}
  \small
  \begin{tabular}{llrrrrr}
    \toprule
    Mechanism & $\varepsilon$ & 10K & 25K & 50K & 250K & 500K \\
    \midrule
    PrivBayes & 0.1 & 0.310 & 0.241 & 0.282 & 0.103 & 0.087 \\
     & 1 & 0.186 & 0.135 & 0.141 & 0.087 & 0.051 \\
     & 5 & 0.099 & 0.085 & 0.091 & 0.085 & 0.046 \\
     & 10 & 0.092 & 0.083 & 0.078 & 0.083 & 0.046 \\
     & 20 & 0.070 & 0.073 & 0.078 & 0.082 & 0.098 \\
    \midrule
    MST & 0.1 & 0.536 & 0.524 & 0.382 & 0.169 & 0.120 \\
     & 1 & 0.237 & 0.135 & 0.106 & 0.079 & 0.082 \\
     & 5 & 0.101 & 0.081 & 0.078 & 0.070 & 0.070 \\
     & 10 & 0.072 & 0.074 & 0.064 & 0.066 & 0.069 \\
     & 20 & 0.080 & 0.071 & 0.061 & 0.067 & 0.070 \\
    \midrule
    SAVANT & 0.1 & 0.623 & 0.528 & 0.545 & 0.690 & 0.697 \\
     & 1 & 0.257 & 0.213 & 0.267 & 0.154 & 0.124 \\
     & 5 & 0.180 & 0.164 & 0.145 & 0.048 & 0.053 \\
     & 10 & 0.172 & 0.162 & 0.080 & 0.065 & 0.035 \\
     & 20 & 0.150 & 0.167 & 0.047 & 0.056 & 0.039 \\
    \bottomrule
  \end{tabular}
\end{table}

\paragraph{For marginal mechanisms, more data buys back the cost of a strict
budget.}
Table~\ref{tab:headline} gives, for each mechanism on ACS-PUMS, the utility drop
at each end of the budget range and the penalty $\Delta$ for moving between
them. MST's penalty falls from $+0.456$ at 10K rows to $+0.050$ at 500K, a
$9\times$ reduction for $50\times$ the data. PrivBayes follows the same broad
direction. Adult shows the same endpoint direction over its narrower range
(PrivBayes $+0.302 \rightarrow +0.216$ from 10K to 48.8K rows). The pattern
holds under two different privacy definitions (PrivBayes is pure
$\varepsilon$-DP, MST is $(\varepsilon,\delta)$-DP), which suggests a family
effect, while the non-monotonic cell argues against calling it a law.

\paragraph{SAVANT performs much better with more data.}
At $\varepsilon=20$ SAVANT converts data into utility more steeply than either
baseline, from 0.150 at 10K rows to 0.039 at 500K, the best result in the
table; at $\varepsilon=1$ it improves from 0.257 to 0.124. The margin over MST
at 500K is about 0.03 F1, close to our seed-noise floor, so we read it as
consistent with neural mechanisms becoming competitive when both budget and
data are abundant~\cite{chen2025benchmarking} rather than as a ranking. At
$\varepsilon=0.1$ the penalty does not shrink: the drop stays between 0.53 and
0.70 at every size, leaving F1 within about 0.05 of zero at 250K and 500K. We
do not read trends inside that column.

\paragraph{Noise scales differently by family.}
A scaling argument predicts the shape. A marginal released over $n$ rows has
magnitude $\Theta(n)$. The Laplace mechanism~\cite{dwork2006calibrating} adds
noise of scale $\Delta/\varepsilon$, and a count's sensitivity $\Delta=1$ does
not grow with $n$, so for a fixed set of queries the relative error is
$\Theta(1/n\varepsilon)$ and iso-utility contours are curves of constant
$n\varepsilon$. For DP-SGD with lot size $L$, sampling rate $q=L/n$, and $T$
steps, the moments accountant~\cite{abadi2016deep} admits a noise multiplier
$\sigma=\Theta(q\sqrt{T\log(1/\delta)}/\varepsilon)$. Gaussian noise of scale
$\sigma C$ on a sum of $L$ gradients clipped to norm $C$ perturbs the averaged
gradient by $\sigma/L=\Theta(\sqrt{T\log(1/\delta)}/n\varepsilon)$. At a fixed
number of epochs $T=\Theta(n/L)$, this is $\Theta(1/\varepsilon\sqrt{n})$, a
strictly worse rate, the direction we observe. All rates hold up to constants,
for a fixed query set and lot size. The argument is asymptotic and assumes gradient norms sit at the
clipping threshold. It holds over the middle of the grid (MST's drops agree within 0.005 F1
at equal $n\varepsilon$) and fails at the strict end, where PrivBayes at 500K rows and $\varepsilon=0.1$ beats the 50K, $\varepsilon=1$
cell with the same $n\varepsilon$.

\begin{figure}[t]
  \centering
  \includegraphics[width=\columnwidth]{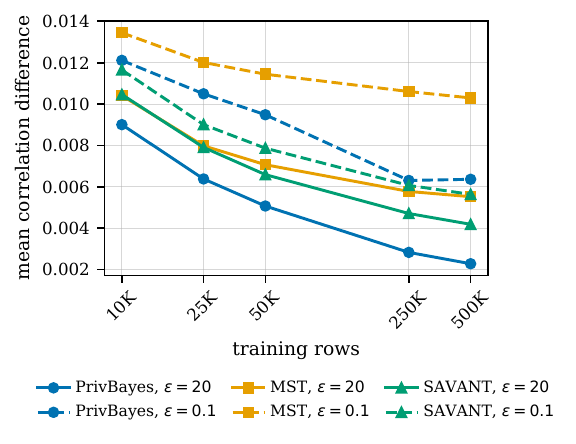}
  \caption{Mean correlation difference over all column pairs on ACS-PUMS at the
  two budget extremes. Lower is better; 0 would preserve every pairwise
  relationship. Solid lines are $\varepsilon=20$, dashed lines $\varepsilon=0.1$.}
  \label{fig:corr}
\end{figure}

\paragraph{Pairwise structure corroborates this and reorders the mechanisms.}
Utility rests on one task and model class, so we check the same axes with the
mean absolute difference in correlation over every column pair, categoricals
one-hot expanded. Figure~\ref{fig:corr} plots it against size on ACS-PUMS: the
error generally falls with more data for every mechanism at both budget
extremes. The data effect appears on Adult too: MST's
$\varepsilon$ 20 to 0.1 correlation penalty falls from $+0.041$ at 10K rows to
$+0.008$ at 48.8K. The metric is informative where utility is not:
SAVANT's $\varepsilon=0.1$ correlation error improves steadily with size, from
0.0117 at 10K rows to 0.0056 at 500K on ACS, even though its utility drop does
not. More data was still buying distributional structure at that budget; it
did not buy a table a classifier could learn the label from. 

\paragraph{Cost.}
SAVANT requires substantially more compute than the marginal mechanisms, which
may matter more than a 0.03 F1 edge for pipelines that regenerate a synthetic
table regularly.

\section{Empirical Privacy Audit}
\label{sec:audit}

The $(\varepsilon,\delta)$ guarantee bounds worst-case leakage by construction.
We add two empirical probes on every cell, each judged against a measured
reference point rather than against zero.

\paragraph{Membership inference.}
A distance-to-closest-record attack~\cite{houssiau2022tapas,yao2025dcr} scores
each probe record by the Gower distance~\cite{gower1971general} to its nearest
synthetic row; members should sit closer than held-out non-members if
membership leaks. Across 222 cells the mean ROC-AUC is \textbf{0.4992}
(sd 0.0073, 95\% CI [0.4982, 0.5001]), indistinguishable from chance. The same
attack on realistic but provably independent real data gives a null with mean
0.4989 (range 0.484 to 0.510), so measured and null distributions share centre,
spread, and range. DOMIAS~\cite{vanbreugel2023domias} looks for local
overfitting instead of proximity: it fits one density model to the synthetic
data and another to a reference sample of real records, and scores a probe as a
member when the synthetic density at that record is high relative to the
reference density, as it would be if the generator had memorized that region of
its training data. It finds the same (mean AUC about 0.50), and the
per-$\varepsilon$ means show no
upward trend as noise shrinks.

\paragraph{Copy detection.}
Exact-match rates are 0.000\% for PrivBayes and MST and <0.1\% for
SAVANT at ACS 250K and 500K. The reference is how often \emph{unseen} real
people match a training row on this schema, 1.4\% and 2.5\%; every mechanism
sits about $30\times$ below it, so matches reflect population density rather
than memorization.

\paragraph{Scope of the negative result.}
Both probes are the weak class: aggregate and distance- or density-based.
Critiques show such proxies can pass data that per-record shadow-model attacks
break~\cite{yao2025dcr,ganev2025inadequacy}, though those studies evaluate
synthesizers with no formal guarantee. At $\varepsilon\geq 10$ the formal
membership-advantage bound $(e^{\varepsilon}-1)/(e^{\varepsilon}+1)$~\cite{yeom2018privacy}
is about 1.0, so our null result shows this attack did not approach the
worst case, not that the mechanisms resist stronger ones. The $(\varepsilon,\delta)$
guarantee, not these probes, bounds a theoretical worst case adversary.

\section{Conclusion}

For marginal mechanisms, a larger table absorbed most of what a strict budget
cost, and the exchange rate between data and budget differs for DP-SGD. SAVANT,
a self-attentive DP VAE, follows the same direction at $\varepsilon\geq 1$ and
is the strongest mechanism at large data and loose budget, but does not train
reliably at $\varepsilon=0.1$. Two independent membership-inference attacks
found no aggregate signal against a measured null, with the caveat that both
are weak. We offer these as evidence from one case with public stand-in data,
for teams deciding how much data to feed a private generator.

\bibliographystyle{ACM-Reference-Format}
\bibliography{references}

\end{document}